\documentclass[runningheads]{llncs}

\usepackage{accv}

\usepackage{accvabbrv}

\usepackage{booktabs}
\usepackage{graphicx}
\usepackage{pifont}
\newcommand{\cmark}{\ding{51}}
\newcommand{\xmark}{\ding{55}}

\ifdefined\pdfcompresslevel
\usepackage[accsupp]{axessibility}
\fi  %

\usepackage[pagebackref,breaklinks,colorlinks,citecolor=accvblue]{hyperref}

\usepackage{orcidlink}

\begin{document}

\title{CrossDD: Cross-Reference Disparity Estimation and Defocus Deblurring Using Dual-Pixel Sensor}
\titlerunning{CrossDD}

\author{Zhuofeng~Wu\orcidlink{0009-0005-4690-7403} \and
Yusuke~Monno\orcidlink{0000-0001-6733-3406} \and
Masatoshi~Okutomi\orcidlink{0000-0001-5787-0742}}

\authorrunning{Z.~Wu et al.}

\institute{
Institute of Science Tokyo, Tokyo, Japan\\
\email{zwu@ok.sc.e.titech.ac.jp}
}

\maketitle

\begin{abstract}
Dual-Pixel (DP) sensors provide two sub-aperture views in a single capture, offering tiny-baseline stereo disparities for geometry estimation and dual-view defocus observations for image restoration. 
This property makes DP sensing attractive for disparity estimation and defocus deblurring.
Although these two tasks are complementary, i.e., a DP disparity map provides blur-amount cues for deblurring and a deblurred image provides sharp object boundaries for disparity edges, existing studies rarely address these tasks jointly.
In this paper, we propose CrossDD, a unified network architecture for joint DP disparity estimation and defocus deblurring. 
Built on a recurrent DP disparity estimation backbone, CrossDD introduces a Cross-Reference Refinement (CRR) module after each recurrent update. 
At every iteration, the CRR module refines the current disparity and image in a cross-reference manner, where the disparity is utilized as the guide for refining the image and vice versa, enabling bidirectional interaction between disparity estimation and defocus deblurring throughout the recurrent process.
Since existing real-world DP image datasets provide geometric ground truths in the form of depth maps, we convert them to disparity maps by estimating a blur scale parameter, effectively repurposing the dataset for our joint task. 
Experimental results on both synthetic and real-world datasets demonstrate that our proposed CrossDD achieves state-of-the-art performance in both disparity estimation and defocus deblurring.

\end{abstract}

\section{Introduction}
\label{sec:introduction}

Dual-Pixel (DP) sensors have become widely available in modern cameras and smartphones, where each pixel is split into two sub-aperture views~\cite{garg2019learning}.
Compared with conventional image sensors and stereo cameras, DP sensors have the advantage that they capture a DP image pair, i.e., left and right images with disparities, in one shot using a single camera. Since the DP disparity is related to the amount of blur, i.e., the distance from the focus plane of the camera, it can be useful for various computer vision and computational photography applications, such as depth/disparity estimation~\cite{garg2019learning}, defocus deblurring~\cite{abuolaim2020defocus}, autofocusing~\cite{choi2023exploring}, reflection removal~\cite{punnappurath2019reflection}, and raindrop removal~\cite{li2024dual}.

DP depth/disparity estimation is a promising topic for geometry estimation using a single camera.
The DP disparity is convertible to the scene depth with an affine transformation, given the camera parameters~\cite{garg2019learning}. Thus, existing methods estimate depth or disparity from a DP left-right image pair through stereo matching~\cite{monin2024continuous} or learning-based regression~\cite{Kim2023Dualpixel}. 
Defocus deblurring is another well-studied topic for DP sensors. DP deblurring methods estimate an all-in-focus image from dual-view left and right defocus images with distinct sub-aperture blur characteristics~\cite{abuolaim2020defocus}, which is advantageous over standard single-image deblurring. 

Although most existing DP-based methods address defocus deblurring and depth/disparity estimation as separate problems, they are tightly related.
Regarding deblurring, the DP disparity for each pixel provides a direct physical cue of the blur amount~\cite{punnappurath2020modeling}. Given this, several methods incorporate disparity-related cues into their network architectures~\cite{liang2022bambnet,li2025dpanet,yang2023k3dn}. 
Regarding depth/disparity estimation, an RGB image is useful to refine depth/disparity edges because RGB object boundaries are well correlated to depth/disparity edges~\cite{ferstl2013image}.
Given this, some DP-based methods apply RGB-guided disparity refinement~\cite{kurita2025revisiting}. Despite these studies, jointly addressing defocus deblurring and depth/disparity estimation has rarely been explored.

To our knowledge, only the study by Pan et al. explicitly addresses the joint task of depth estimation and defocus deblurring using a DP sensor~\cite{pan2021dual, pan2024weakly}. They introduced a reblur loss, which takes the differences between the input DP image pair and the simulated DP image pair from the estimated all-in-focus image and depth map through a DP image formation model. This enables joint end-to-end training of depth estimation and deblurring and enhances the consistency between their results. On the other hand, their estimation networks are separated and sequential, not including the explicit bidirectional interaction between the two tasks at the network level.

\begin{figure}[t]
    \centering
    \includegraphics[width=0.8\linewidth]{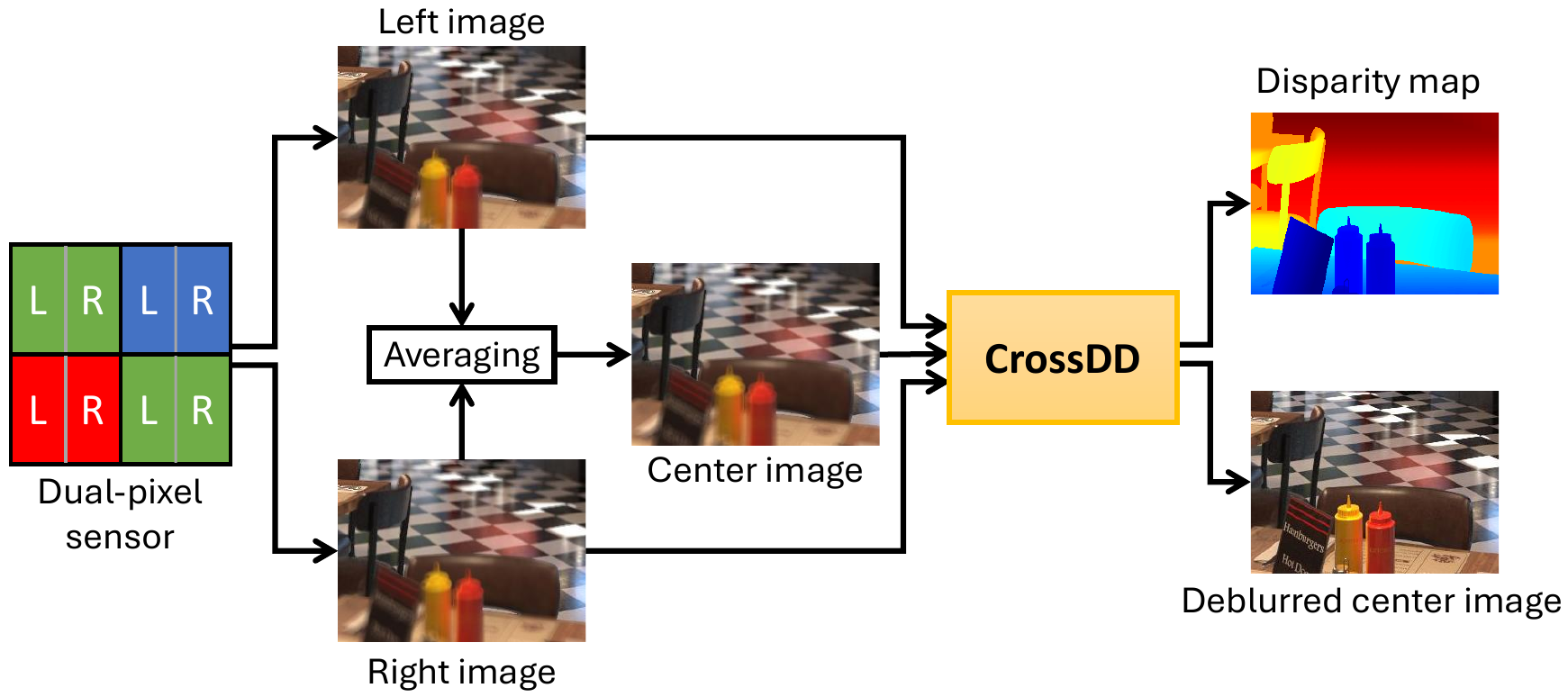}\\ \vspace{-3mm}
    \caption{
    Overview of the network inputs and outputs of our proposed CrossDD. A dual-pixel sensor provides left and right sub-aperture images, whose average forms the center image.
    Given the left, right, and center images, CrossDD jointly estimates the center-referenced disparity map and the corresponding deblurred image.
    }
    \label{fig:overview}
    \vspace{-5mm}
\end{figure}

In this paper, we propose CrossDD, a unified network architecture for disparity estimation and defocus deblurring from a DP image pair.
Figure~\ref{fig:overview} provides an overview of the network inputs and outputs of our proposed CrossDD.
From the raw DP sensor output, the left and right sub-aperture images are obtained, and their average is used as the center image.
CrossDD takes the left, right, and center images as inputs and jointly estimates the center-referenced disparity map and the corresponding deblurred image.
The network architecture is built on a recurrent disparity estimation backbone and introduces a key module named Cross-Reference Refinement~(CRR) module after each recurrent update.
At every iteration, the current disparity estimate is utilized to refine the current deblurred image and vice versa.
In this way, the disparity provides geometry-aware guidance for deblurring, while the deblurred image provides clearer structures and boundaries for disparity refinement.
Unlike prior joint methods applying sequential estimations in one direction and in one-path~\cite{pan2021dual, pan2024weakly}, CrossDD enables bidirectional and iterative interactions between disparity estimation and defocus deblurring throughout the recurrent process.

The training of CrossDD requires triplets of input DP images, ground-truth disparity maps, and all-in-focus images. Although these triplets are publicly available in the synthetic dataset of~\cite{Wu_2024_BMVC}, no real-world dataset is currently available. This is because existing DP image datasets provide the geometric ground truths in the form of depth maps, not disparity maps, mainly for two practical reasons. (i)~The acquisition of depth maps and their alignment to the camera coordinates are relatively well-established, such as in structured-light techniques and time-of-flight sensing. (ii)~The acquisition of ground-truth DP disparities requires the calibration of blur size according to scene depth, for which a reliable method has not yet been established. Given these, we propose a method to convert the depth maps to the DP disparity maps by estimating a blur size parameter and apply this method to an existing real-world DP image dataset to repurpose it for our joint task.  

The main contributions of this work are summarized as follows.
\begin{itemize}
    \item We propose CrossDD, a joint network architecture for DP disparity estimation and defocus deblurring. We propose a key module named the CRR module, which couples the two tasks in a recurrent and bidirectional manner to improve each other.
    \item We propose a method to convert ground-truth depth maps to DP disparity maps through blur size parameter estimation. Using this method, an existing DP dataset is repurposed to train the joint task.
    \item We demonstrate through experiments on both synthetic and real datasets that our CrossDD achieves strong performance on both disparity estimation and defocus deblurring.
    \item We make our dataset and source code publicly available\footnote{\url{http://www.ok.sc.e.titech.ac.jp/res/DPQP/}}.
\end{itemize}

\section{Related Work}

\subsection{Dual-Pixel Depth/Disparity Estimation and Defocus Deblurring}

DP depth/disparity estimation methods can be roughly classified into optical modeling-based methods~\cite{punnappurath2020modeling}, stereo matching methods~\cite{monin2024continuous}, and learning-based methods~\cite{garg2019learning,Kim2023Dualpixel,kurita2025revisiting}. 
Punnappurath et al. explicitly modeled the defocus-disparity relationship in DP sensors with a parametric point spread function for depth estimation~\cite{punnappurath2020modeling}. 
Monin et al. introduced continuous cost aggregation within a semi-global matching framework to estimate continuous DP disparities~\cite{monin2024continuous}. 
Learning-based methods have been actively studied. Garg et al. proposed a supervised method considering the affine ambiguity between scene depth and DP disparity~\cite{garg2019learning}, Kim et al. proposed a self-supervised framework based on the reflective symmetry of anisotropic DP blur kernels~\cite{Kim2023Dualpixel}, and Kurita et al. combined classical stereo matching with learning-based disparity completion~\cite{kurita2025revisiting}. 
Recent works further extended disparity estimation to quad-pixel~(QP) sensors, which observe disparities in both horizontal and vertical directions~\cite{Wu_2024_BMVC,yu2025all}.

DP defocus deblurring has also been studied using both optical modeling and learning-based approaches. 
Xin et al. proposed an optimization method that jointly estimates a defocus map and an all-in-focus image from a DP image pair by explicitly modeling DP optics~\cite{Xin2021dualpixel}. 
Abuolaim and Brown introduced a learning-based DP deblurring network and demonstrated the advantage of using DP sub-aperture views for defocus deblurring~\cite{abuolaim2020defocus}. 
Subsequent methods improved DP deblurring by blur-aware multi-branch processing~\cite{liang2022bambnet}, disparity-aware kernel estimation and reblurring regularization~\cite{yang2023k3dn}, CLIP-based language supervision for blur-map estimation~\cite{yang2024ldp}, and explicit alignment between DP views using deformable convolutions~\cite{li2025dpanet}. 
A recent work also applied QP sensing to defocus deblurring~\cite{chen2025quad}.

Despite the effectiveness of each method, these methods focus on either disparity estimation or defocus deblurring, not jointly handling these two tasks.

\subsection{Joint Approach}

The methods by Pan et al. jointly estimate the depth and restore the all-in-focus image from a DP image pair~\cite{pan2021dual,pan2024weakly}.
These methods first apply a depth estimation network to an input DP image pair and then apply a deblurring network, where the estimated depth is used as an additional input to the DP image pair. 
Those networks are jointly trained in an end-to-end manner using a reblur loss that enhances the consistency between the estimated depth map and all-in-focus image through a DP image formation model. 
In addition to the reblur loss, the first method~\cite{pan2021dual} applies supervised losses for the depth and the all-in-focus image, while the extended method~\cite{pan2024weakly} removes the necessity of the ground-truth depth by using focal stacks for self-supervision.

Our CrossDD differs from these joint methods mainly in that it enforces bidirectional and recurrent interactions between disparity estimation and deblurring, while the existing methods include only one-directional and one-path interaction in their network architectures.

\subsection{Dual-Pixel Image Datasets}

For joint DP disparity estimation and defocus deblurring, suitable datasets are still limited. 
For synthetic data, QPD-Data~\cite{Wu_2024_BMVC} provides both disparity ground truths and all-in-focus images because all necessary data can be generated by simulation.
In contrast, real-world DP datasets are usually designed for either geometry estimation or image restoration. 
Geometry-oriented datasets provide depth- or disparity-related supervision but lack paired all-in-focus images~\cite{garg2019learning,punnappurath2020modeling,kang2022facial}, while deblurring-oriented datasets provide all-in-focus images but lack paired disparity ground truths~\cite{abuolaim2020defocus,abuolaim2022improving}.
Among existing real-world DP datasets, DP5K~\cite{li2023learning} is the closest to our target setting because it provides real DP image pairs, sharp images, camera parameters, and depth maps. 
We therefore repurpose DP5K for our joint task through depth-to-disparity conversion.
\section{Proposed Method: CrossDD}
\subsection{Overall Network Architecture}

\begin{figure*}[t!]
    \centering
    \includegraphics[width=\textwidth]{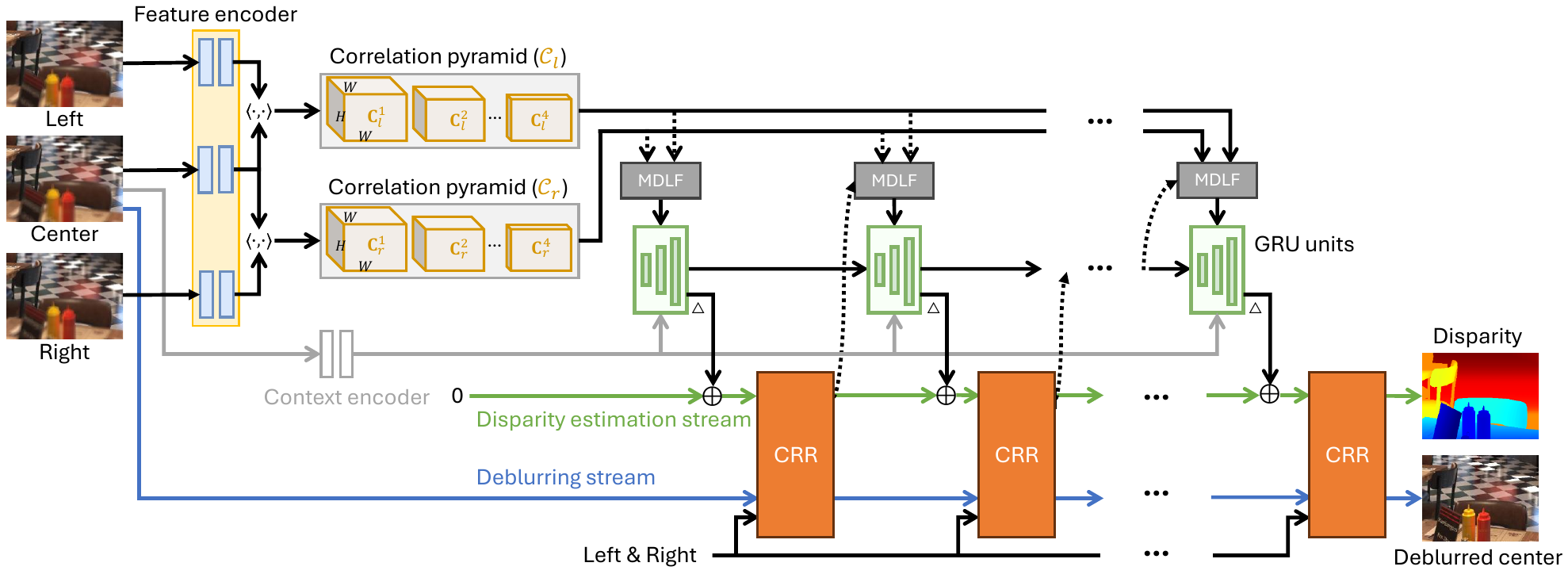}\\   \vspace{-2mm}
    \caption{
    Architecture of the proposed CrossDD. 
    Given DP image inputs, the network recurrently estimates the disparity map and the deblurred image. We introduce the proposed CRR module after each recurrent update for joint refinement of the disparity map and the deblurred image. 
    }
    \label{fig:architecture}
    \vspace{-3mm}
\end{figure*}

Figure~\ref{fig:architecture} shows the overall network architecture of our proposed CrossDD. 
Given the triplet of the left, center, and right images, the network predicts a disparity map and a deblurred image, both defined in the center-image reference frame. 

The upper part of Fig.~\ref{fig:architecture} shows the disparity estimation backbone, for which we adopt the architecture of QPDNet~\cite{Wu_2024_BMVC}. Although QPDNet originally targets disparity estimation for QP sensors, we apply it for DP sensors by removing the disparity estimation from the vertical disparities. 
The features are first extracted from the left, center, and right images, respectively, using a shared encoder. The bidirectional (center-left and center-right) correlation pyramids are then constructed with the center image as a reference by the inner product of the features shifted in the horizontal direction. 
At each iteration, local correlation features are retrieved according to the current disparity estimate using a Multi-Directional Lookup Fusion (MDLF) strategy proposed in~\cite{Wu_2024_BMVC}.
They are fed into the Gated Recurrent Unit (GRU) together with the current disparity (initialized as zero) and the context features extracted from the center image to progressively update the disparity estimate.

To explicitly couple disparity estimation and deblurring, our proposed CRR module is inserted after every GRU update. Deblurring is performed inside the CRR module.
At each iteration, the current disparity from the GRU is passed to the CRR module together with the current deblurred image (initialized by the center image) and the original left and right images. 
The CRR module then performs sequential cross-task refinement, where the disparity provides geometry-aware guidance for deblurring and the deblurred image provides clearer structural cues for disparity refinement.

\subsection{Cross-Reference Refinement Module}
\label{sec:crm}

\begin{figure}[t]
    \centering
    \includegraphics[width=\linewidth]{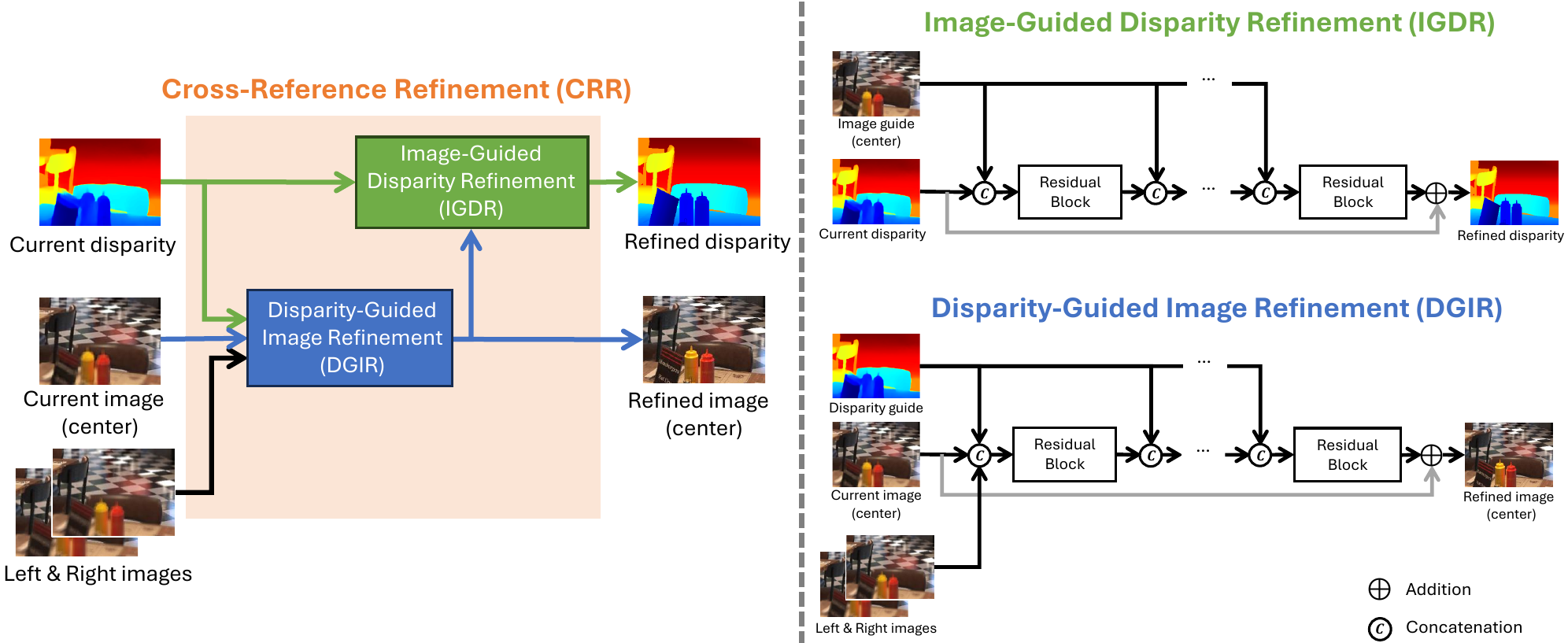}\\    \vspace{-3mm}
    \caption{
    Left: Overview of the proposed CRR module.
    DGIR refines the current center image using the disparity as geometry guidance. IGDR refines the current disparity using the center image as structural guidance.
    Right: IGDR and DGIR consist of a stack of residual blocks to refine the disparity and the image, respectively.
    }
    \label{fig:crm}
    \vspace{-3mm}

\end{figure}

As shown in the left side of Fig.~\ref{fig:crm}, the CRR module consists of two submodules: Disparity-Guided Image Refinement (DGIR) module and Image-Guided Disparity Refinement (IGDR) module. Given the current disparity map from the $j$-th GRU,~$\mathbf{D}_j$, the current deblurred image, $\mathbf{S}_j$, and the original left and right images, ($\mathbf{I}_L,\mathbf{I}_R$), the CRR module first refines the current image through the DGIR module using disparity-aware guidance, and then refines the current disparity through the IGDR module using the refined image as structural guidance (The order of these modules can be reversed technically). 
The sequential update process is formulated as
\begin{equation}
\tilde{\mathbf{S}}_j = \mathrm{DGIR}\!\left(\mathbf{S}_j, \mathbf{D}_j, \mathbf{I}_L, \mathbf{I}_R\right), \ \ 
\tilde{\mathbf{D}}_j = \mathrm{IGDR}\!\left(\mathbf{D}_j, \tilde{\mathbf{S}}_j\right),
\end{equation}
where $\tilde{\mathbf{S}}_j$ and $\tilde{\mathbf{D}}_j$ denote the refined image and disparity, respectively. These refined estimates are then passed to the next iteration.

\subsubsection{Disparity-Guided Image Refinement}

The DGIR module aims to refine the current image with the guidance of the disparity. 
As shown in the lower right of Fig.~\ref{fig:crm}, the DGIR module takes the current image, the disparity guide, and the original left and right images as the inputs. 
These inputs are concatenated and fed into a stack of $M$ residual blocks~\cite{yu2018wide}. Each residual block consists of two $3\times3$ convolutions with a ReLU activation between them, and outputs a residual feature that is added to the input feature through a skip connection.
In addition, the disparity guide is repeatedly concatenated with the intermediate feature after each residual block and then passed to the next block, allowing the geometric guidance to be injected throughout the refinement process. 
In particular, the disparity provides explicit depth-related guidance, while the left and right images provide the original sub-aperture observations and complementary phase and blur cues that are not explicitly retained in the current image.
The current image is initialized from the center image as the spatial reference, which avoids the misalignment between the disparity and the image.

\subsubsection{Image-Guided Disparity Refinement}

The IGDR module aims to refine the current disparity map using the structural information provided by the deblurred image. 
As illustrated in the upper right of Fig.~\ref{fig:crm}, the IGDR module shares the same residual refinement architecture as the DGIR module, including the same number of residual blocks, but targets disparity refinement instead of image refinement.
It takes the current disparity map and the image guide as the inputs, where the image guide is concatenated with the intermediate feature after each residual block. 
In this manner, the IGDR module helps recover more accurate disparity boundaries and finer geometric details through the refinement process.

\subsection{Loss Function}
\label{sec:loss}

We train CrossDD in two stages for stable optimization. 
In the first stage, all CRR modules are removed, and only the recurrent disparity estimation backbone is trained. 
The first-stage loss is therefore defined only on the disparity predictions from the GRU-based backbone as
\begin{equation}
\mathcal{L}_{\mathrm{stage1}}
=
\sum_{j=1}^{N}
\gamma^{\,N-j}
\left\|
\hat{\mathbf{D}}-\mathbf{D}_j
\right\|_1 ,
\end{equation}
where $\mathbf{D}_j$ is the estimated disparity map from $j$-th GRU and  $\hat{\mathbf{D}}$ is the ground-truth disparity map.
$N$ denotes the total number of recurrent iterations, and $\gamma$ is the decay factor for intermediate supervision.

In the second stage, the disparity estimation backbone is initialized with the trained weights at the first stage and jointly updated with all CRR modules enabled. 
In this stage, the disparity predictions from the GRUs, the refined disparity predictions from the CRR modules, and the refined image predictions from the CRR modules are all supervised. 
The overall loss is defined as
\begin{equation}
\label{formula:loss}
\mathcal{L}_{\mathrm{stage2}}
=
\sum_{j=1}^{N}
\gamma^{\,N-j}
\left(
\left\|
\hat{\mathbf{D}}-\mathbf{D}_j
\right\|_1
+
\lambda_{\mathrm{d}}
\left\|
\hat{\mathbf{D}}-\tilde{\mathbf{D}}_j
\right\|_1
+
\lambda_{\mathrm{s}}
\left\|
\hat{\mathbf{S}}-\tilde{\mathbf{S}}_j
\right\|_1
\right),
\end{equation}
where $\hat{\mathbf{D}}$ and $\hat{\mathbf{S}}$ denote the ground-truth disparity map and all-in-focus image, respectively. 
$\mathbf{D}_j$ and $\tilde{\mathbf{D}}_j$ are the disparity maps from the GRU and the CRR module, respectively, and $\tilde{\mathbf{S}}_j$ is the deblurred image from the CRR module.
Here, $\lambda_d$ and $\lambda_s$ are hyperparameters to balance each term.

\section{Real-World Dataset Construction}
\label{sec:dataset}

\subsection{Review of DP5K Dataset}

Existing real-world DP image datasets provide geometric ground truths as depth maps rather than DP disparity maps for some practical reasons, as mentioned in the introduction. Thus, we converted the depth maps into DP disparity maps for our task. For this purpose, we utilized the DP5K dataset~\cite{li2023learning}, which is one of the largest datasets including ground-truth depth maps and sharp images. 

The ground-truth depth maps of DP5K were obtained using a structured-light technique with one projector and two DP cameras in a stereo setting. For each DP camera, DP image pairs were captured under five F-number settings (F1.8, F2.0, F2.8, F4.0, and F5.6), and a sharp image was captured with a large depth-of-field setting of F22. We used this F22 image as the ground truth for defocus deblurring. We confirmed that the ground-truth depth maps and sharp images are well aligned, making them suitable for constructing a real-world benchmark for joint DP disparity estimation and defocus deblurring. 

Since the DP images and sharp images are captured under different F-number settings, they contain brightness differences and vignetting inconsistencies. To reduce their influence on our dataset construction and training, we applied photometric alignment to the DP images using global brightness scaling and radial vignetting correction. Details of these processes are provided in the supplementary material.

\subsection{Depth-to-Disparity Conversion}

\begin{figure}[t]
\begin{center}
\centerline{\includegraphics[width=0.9\linewidth]{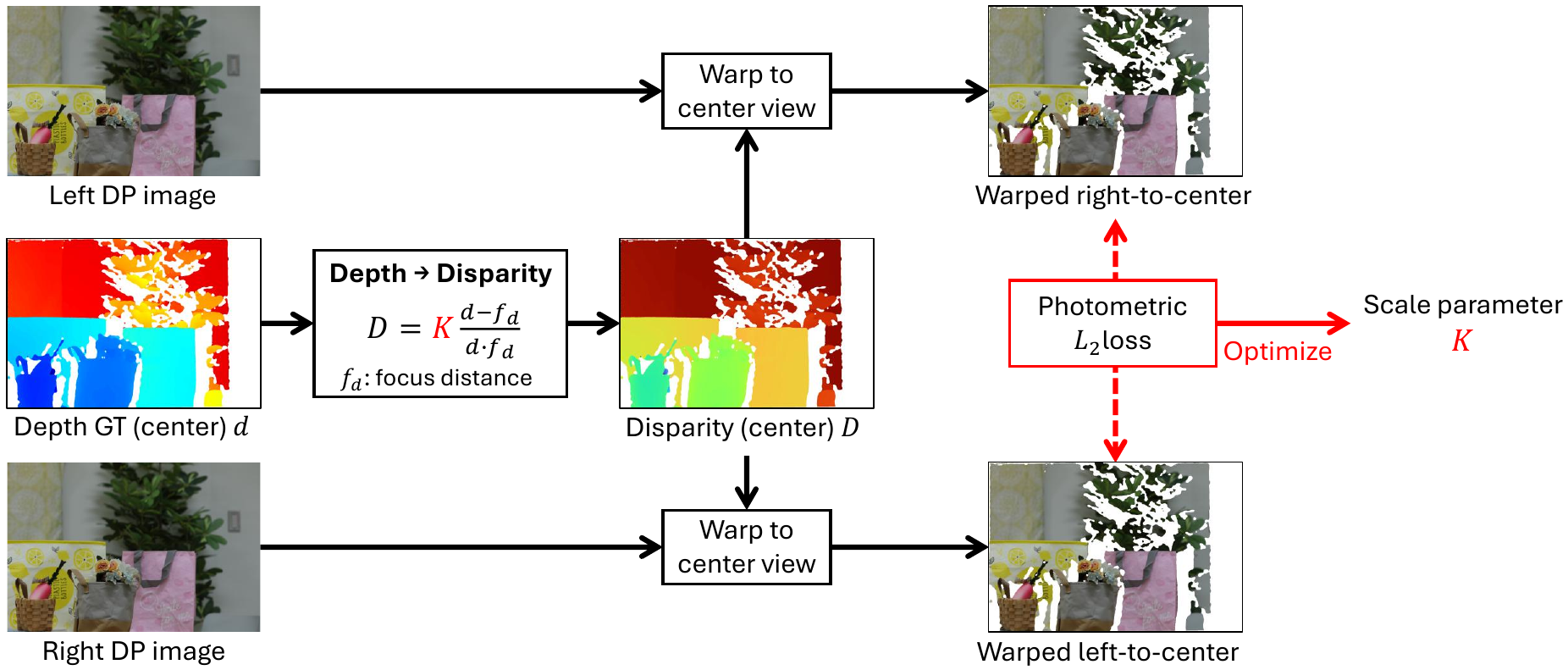}}
\end{center}
   \vspace{-10mm}
   \caption{Pipeline for estimating the blur scale parameter $K$ for depth-to-disparity conversion. The parameter $K$ is used to convert depth into disparity and warp the left and right DP views toward the center view. The optimal $K$ is obtained by minimizing the photometric $L_2$ loss between the warped images.}
\label{fig:k_estimation}
\vspace{-5mm}
\end{figure}

To convert the depth maps to DP disparity maps, we adopted the thin-lens DP disparity model~\cite{Kim2023Dualpixel}. 
In this model, DP disparity is proportional to the defocus-induced circle of confusion and can be written as
\begin{equation}
\begin{aligned}
D
= \alpha \frac{Af}{1-f/f_d}\left(\frac{1}{f_d}-\frac{1}{d}\right)
= \alpha Af \frac{d-f_d}{d(f_d-f)}
\approx K\frac{d-f_d}{d \cdot f_d},
\end{aligned}
\label{eq:depth_to_disp_signed}
\end{equation}
where $D$ denotes the signed DP disparity, $A$ is the aperture diameter, $f$ is the focal length, $f_d$ is the focus distance, $d$ is the depth value, and $\alpha$ is an ambiguous scale coefficient. 
Since $f \ll f_d$ in our setting, we approximated $f_d-f \approx f_d$ and absorbed the camera-dependent and fixed optical factors into a single blur scale parameter $K$.
Since the blur scale parameter $K$ is not provided in DP5K, we estimated it based on photometric consistency between the DP views, as illustrated in Fig.~\ref{fig:k_estimation}. 
Given a candidate $K$ value, we converted the depth map into a disparity map using Eq.~(\ref{eq:depth_to_disp_signed}) and warped the left and right DP images toward the center view. 
The optimal $K$ value was selected by minimizing the $L_2$ photometric error between the warped left-to-center and right-to-center images using a bounded Brent method~\cite{brent2013algorithms}. 
More details of the optimization procedure are provided in the supplementary material.

\begin{figure}[t]
\centering
\includegraphics[width=\textwidth]{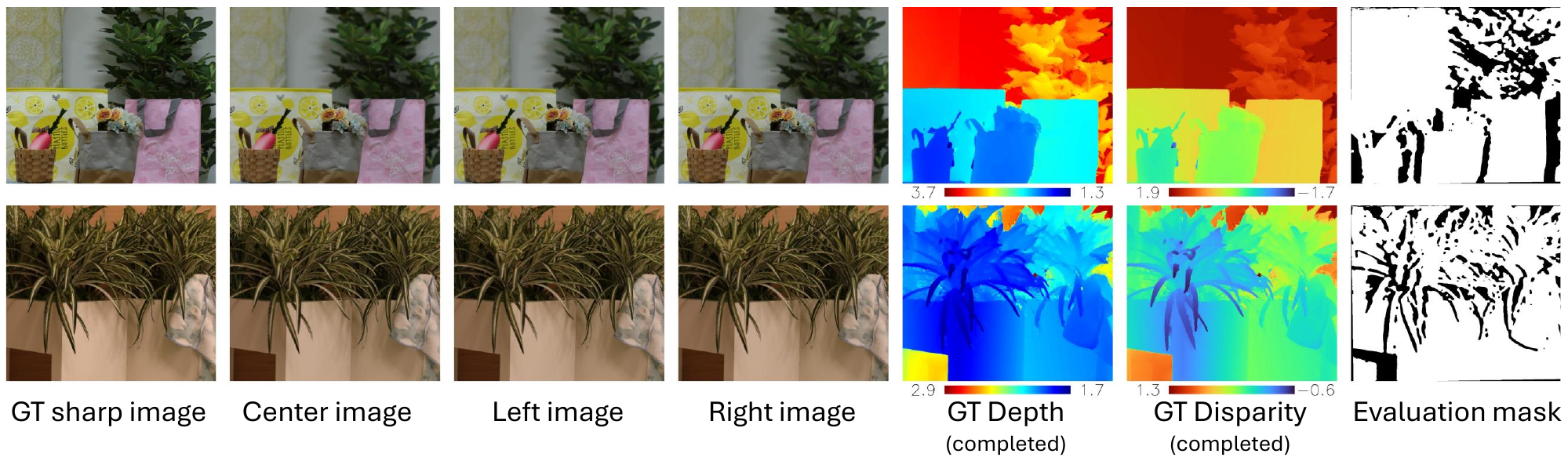}\\ 
\vspace{-2mm}
  \caption{Examples of the resulting DP5K-Disp dataset. From left to right: sharp image, center image, left image, right image, completed depth ground truth, generated disparity ground truth, and evaluation masks. White and black in the evaluation mask indicate valid and invalid pixels for quantitative evaluation, respectively.}
\label{fig:dpdad_overview}
\vspace{-3mm}
\end{figure}

The original DP5K depth maps contain missing or occluded regions. 
To provide dense disparity supervision, we performed depth completion using Marigold-DC~\cite{viola2025marigold}, while preserving the original depth values in valid regions. 
For quantitative evaluation, we constructed an evaluation mask from the original non-occluded regions. 
In addition, because the right side of each image is outside the reliable structured-light stereo estimation region, we cropped the original $1680 \times 1120$ images to $1344 \times 1120$. 
Figure~\ref{fig:dpdad_overview} shows examples of the resulting dataset, named DP5K-Disp, including DP images, completed depth, converted disparity, and evaluation masks.

\section{Experimental Results}

\subsection{Experimental Settings}
\subsubsection{Datasets}
We used two datasets for training and evaluation. 
The first dataset is QPD-Data, a synthetic quad-pixel dataset provided in~\cite{Wu_2024_BMVC}. 
Each sample has a resolution of $1024 \times 768$ and provides left, right, top, and bottom views, together with a sharp image and disparity ground truth. 
Following~\cite{Wu_2024_BMVC}, we used 3,010, 380, and 380 samples for training, validation, and testing, respectively. 
In our experiments, only the left, center, and right views were used, while the top and bottom views were discarded.
The second dataset is the constructed real-world DP5K-Disp dataset introduced in Sec.~\ref{sec:dataset}. 
Each sample contains the left and right views, a sharp image, disparity ground truth, and an evaluation mask, with an image resolution of $1344 \times 1120$. 
In this study, we used the F4.0 setting for input DP image pairs. The dataset contains 378, 63, and 61 samples for training, validation, and testing, respectively.

\subsubsection{Implementation Details}
We set $\lambda_{\mathrm{d}} = \lambda_{\mathrm{s}} = 1$ and $\gamma=0.9$ in Eq.~\ref{formula:loss}. 
The input images were randomly cropped to $448 \times 448$ for training. 
The learning rate, weight decay, and batch size were set to $2\times10^{-4}$, $1\times10^{-5}$, and 4, respectively. 
We set the number of recurrent iterations to $N=8$ and used $M=8$ residual blocks in both DGIR and IGDR modules.
We adopted a two-stage training strategy introduced in Section~\ref{sec:loss}. 
Using a single NVIDIA H100 GPU, the network was trained for 200,000 iterations in the first stage and 400,000 iterations in the second stage, with a total training time of approximately 120 hours. CrossDD has 13.41M parameters and runs in 146.64 ms for a $896\times672$ input.

\subsection{Comparison with Existing Methods}

We compared the proposed CrossDD with representative methods from three task categories:
1) disparity-estimation methods, including stereo-based methods (RAFT-Stereo~\cite{lipson2021raft} and IGEV-Stereo~\cite{xu2023iterative}) and DP-based methods (CCA~\cite{monin2024continuous}, SFBD~\cite{Kim2023Dualpixel}, and QPDNet~\cite{Wu_2024_BMVC});
2) deblurring methods, DPDNet~\cite{abuolaim2020defocus}, BaMBNet~\cite{liang2022bambnet}, and DPANet~\cite{li2025dpanet};
3) joint disparity-estimation and deblurring method, DDDNet~\cite{pan2021dual}.
Since the official training codes for SFBD and DDDNet are not publicly available, we reimplemented them based on the settings described in their papers. All compared methods were trained using the same datasets for fair comparison.

\begin{table}[t]
\centering
\caption{Comparison on QPD-Data and DP5K-Disp test sets. The asterisk indicates that the training code was implemented by the authors due to its unavailability.
The best and second-best results are highlighted in bold and underlined, respectively.}
\vspace{-2mm}
\label{tab:all_comparison}
\scriptsize
\setlength{\tabcolsep}{2.0pt}
\renewcommand{\arraystretch}{1.05}
\resizebox{\columnwidth}{!}{%
\begin{tabular}{lccccc|ccccc}
\toprule
& \multicolumn{5}{c|}{QPD-Data} 
& \multicolumn{5}{c}{DP5K-Disp} \\
\cmidrule(lr){2-6} \cmidrule(lr){7-11}
& \multicolumn{2}{c}{Disparity} 
& \multicolumn{3}{c|}{Deblur} 
& \multicolumn{2}{c}{Disparity} 
& \multicolumn{3}{c}{Deblur} \\
\cmidrule(lr){2-3} \cmidrule(lr){4-6} 
\cmidrule(lr){7-8} \cmidrule(lr){9-11}
Method 
& EPE$\downarrow$ & RMSE$\downarrow$ 
& PSNR$\uparrow$ & SSIM$\uparrow$ & MAE$\downarrow$
& EPE$\downarrow$ & RMSE$\downarrow$ 
& PSNR$\uparrow$ & SSIM$\uparrow$ & MAE$\downarrow$ \\
\midrule
RAFT-Stereo~\cite{lipson2021raft} 
& 0.030 & 0.082 & -- & -- & -- 
& 0.070 & 0.112 & -- & -- & -- \\

IGEV-Stereo~\cite{xu2023iterative} 
& 0.035 & 0.093 & -- & -- & -- 
& 0.070 & 0.114 & -- & -- & -- \\

CCA~\cite{monin2024continuous} 
& 0.148 & 0.283 & -- & -- & -- 
& 0.346 & 0.702 & -- & -- & -- \\

SFBD$^\ast$~\cite{Kim2023Dualpixel} 
& 0.038 & 0.101 & -- & -- & -- 
& 0.074 & 0.120 & -- & -- & -- \\

QPDNet~\cite{Wu_2024_BMVC} 
& \underline{0.027} & \underline{0.077} & -- & -- & -- 
& \underline{0.069} & \underline{0.110} & -- & -- & -- \\

DPDNet~\cite{abuolaim2020defocus} 
& -- & -- & 27.35 & 0.838 & 0.0259 
& -- & -- & 31.70 & 0.909 & 0.0171 \\

BaMBNet~\cite{liang2022bambnet} 
& -- & -- & 32.37 & 0.940 & 0.0136 
& -- & -- & 34.32 & 0.951 & 0.0132 \\

DPANet~\cite{li2025dpanet} 
& -- & -- & \underline{32.72} & \underline{0.947} & \underline{0.0130} 
& -- & -- & \underline{34.63} & \underline{0.954} & \underline{0.0128} \\

 DDDNet$^\ast$~\cite{pan2021dual} 
& 0.054 & 0.104 
& 28.88 & 0.898 & 0.0238 
& 0.171 & 0.300 
& 31.72 & 0.910 & 0.0176 \\

CrossDD~(ours) 
& \textbf{0.023} & \textbf{0.068} 
& \textbf{37.23} & \textbf{0.974} & \textbf{0.0075} 
& \textbf{0.064} & \textbf{0.105} 
& \textbf{34.85} & \textbf{0.955} & \textbf{0.0124} \\

\bottomrule
\end{tabular}%
}
\end{table}

\subsubsection{Results on Synthetic QPD-Data}

\begin{figure}[t!]
    \centering
    \includegraphics[width=\textwidth]{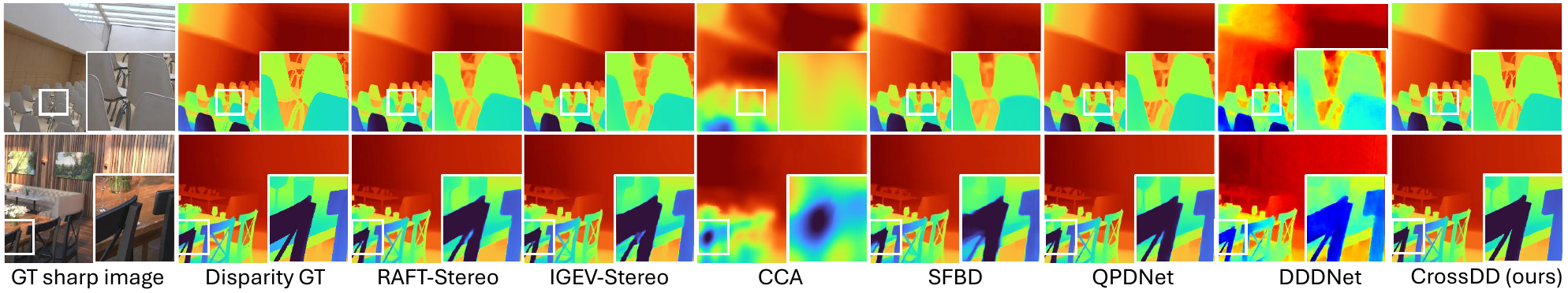}\\
    \vspace{-2mm}
    \caption{
    Qualitative comparison of disparity estimation on the synthetic QPD-Data.
    Our method produces sharper boundaries and more accurate fine structures than the compared methods, especially in the zoomed-in regions indicated by the white boxes.
    }
    \label{fig:syn_disp_vis}
    \vspace{-2mm}
\end{figure}

The quantitative results on the synthetic QPD-Data test set are reported on the left side of Table~\ref{tab:all_comparison}. 
On disparity estimation, CrossDD achieves the best overall accuracy, reaching 0.023 EPE and 0.068 RMSE. 
These results indicate that the proposed recurrent cross-task refinement improves disparity estimation consistently over both stereo-based and DP-based methods on synthetic data.
This tendency is also visible in Fig.~\ref{fig:syn_disp_vis}. 
Compared with the competing methods, CrossDD produces sharper disparity boundaries and more reliable local structures. 
The benefit is especially evident for thin objects and small geometric details, where the competing methods tend to oversmooth the prediction.

\begin{figure}[t!]
    \centering
    \includegraphics[width=\linewidth]{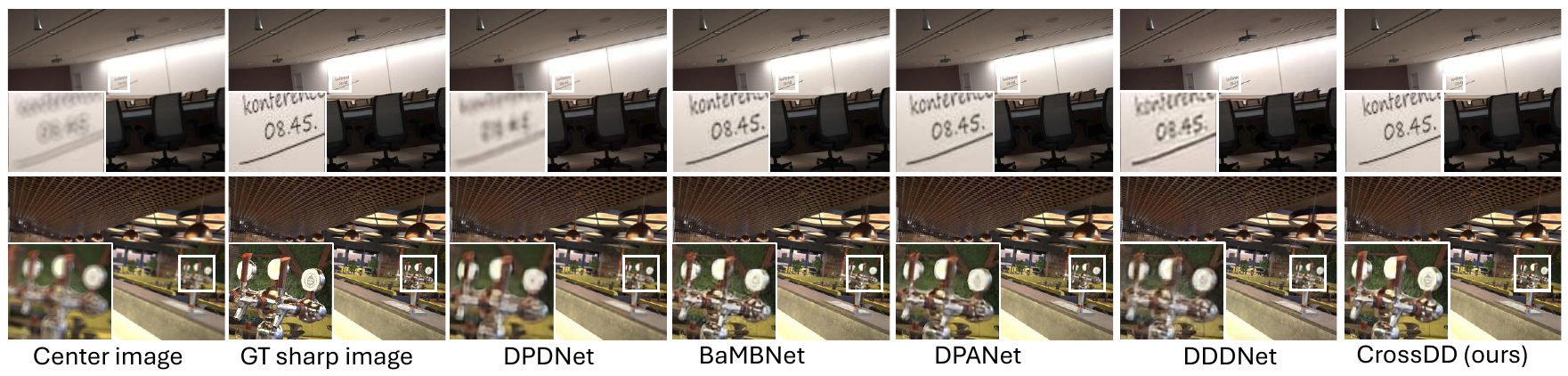}\\ \vspace{-2mm}
    \caption{
    Qualitative comparison of defocus deblurring on the synthetic QPD-Data.
    Our method restores clearer details and sharper local structures than the competing deblurring methods, especially in the zoomed-in regions indicated by the white boxes.
    }
    \label{fig:syn_deblur_vis}
    \vspace{-3mm}
\end{figure}

For image deblurring, CrossDD ranks first on all three metrics. 
Compared with DPANet, the strongest competing deblurring method, it improves PSNR from 32.72 to 37.23, SSIM from 0.947 to 0.974, and reduces MAE from 0.0130 to 0.0075. 
It also surpasses DPDNet and BaMBNet by a large margin, showing that the proposed CRR is highly effective for DP defocus deblurring.
The visual comparisons in Fig.~\ref{fig:syn_deblur_vis} support the same conclusion. 
CrossDD restores clearer textures and sharper fine details, especially in high-frequency regions and small structures, where the competing methods still produce visibly blurred results.
Compared with the joint-task method DDDNet, CrossDD substantially outperforms it on both disparity estimation and defocus deblurring.

\begin{figure}[t]
    \centering
    \includegraphics[width=\textwidth]{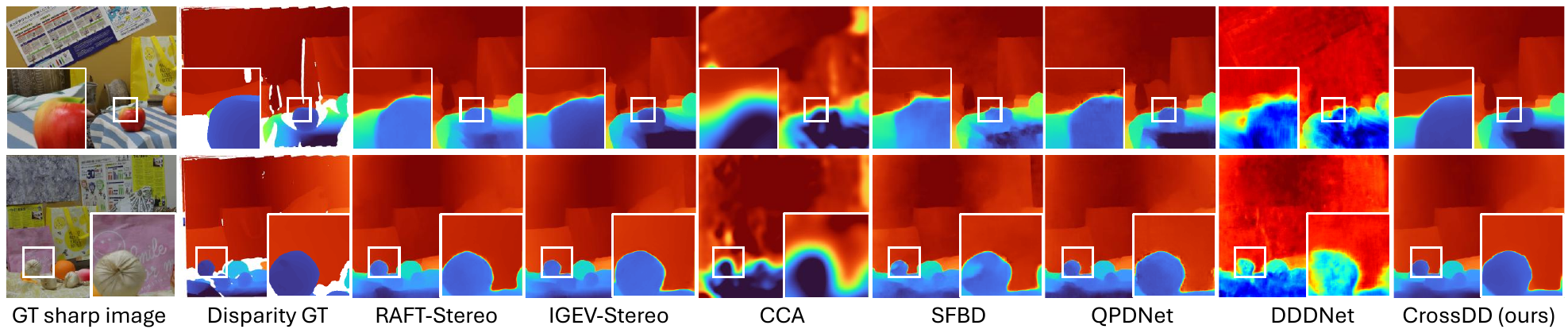}\\ \vspace{-2mm}
    \caption{
    Qualitative comparison of disparity estimation on the real DP5K-Disp.
    Our method yields cleaner disparity maps and better preserves object boundaries than the compared methods.
    }
    \label{fig:real_disp_vis}
    \vspace{-2mm}
\end{figure}

\begin{figure}[t]
    \centering
    \includegraphics[width=\linewidth]{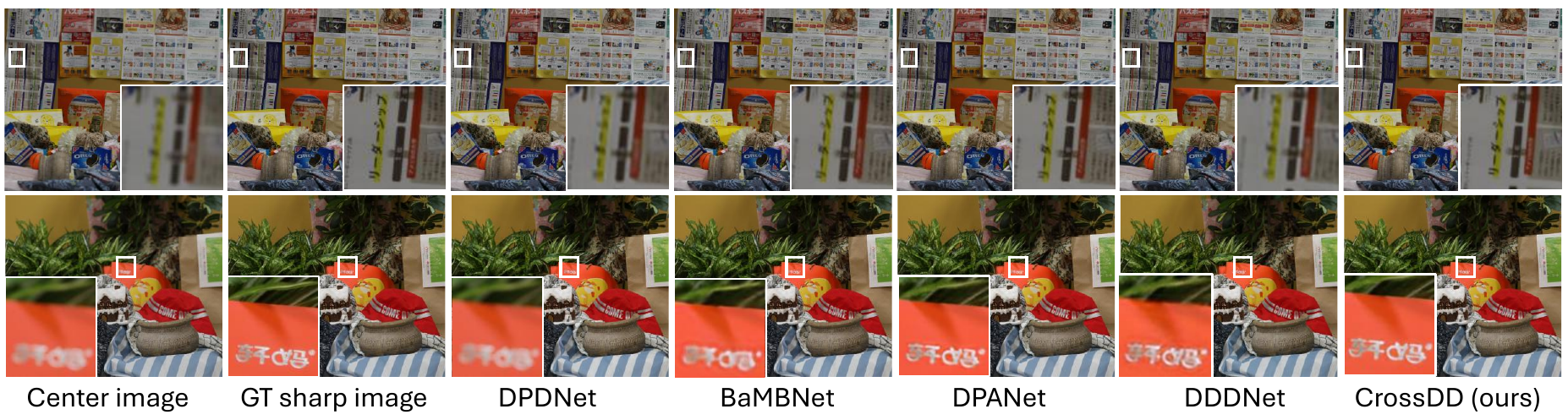}\\ \vspace{-2mm}
    \caption{
    Qualitative comparison of defocus deblurring on the real DP5K-Disp.
    Our method produces sharper restoration results with clearer textures and fewer residual blur artifacts on real-scene DP images.
    }
    \label{fig:real_deblur_vis}
    \vspace{-3mm}
\end{figure}

\subsubsection{Results on Real DP5K-Disp Dataset}

The right side of Table~\ref{tab:all_comparison} summarizes the quantitative results on the real DP5K-Disp test set. 
During the disparity evaluation, we only evaluated the disparity inside the evaluation mask. For disparity estimation, CrossDD achieves the best results among all compared methods, with 0.064 EPE and 0.105 RMSE, showing that the proposed framework is also effective on real-scene DP data.
Figure~\ref{fig:real_disp_vis} shows the same trend qualitatively. 
Compared with the competing methods, CrossDD produces cleaner disparity maps and sharper boundaries, particularly in the zoomed-in regions indicated by the white boxes.

For image deblurring, CrossDD also achieves the best overall performance. 
Compared with the strongest DP deblurring baseline, DPANet, it improves PSNR from 34.63 to 34.85, SSIM from 0.954 to 0.955, and reduces MAE from 0.0128 to 0.0124. 
Although the margin is smaller than on synthetic data, the result still confirms the effectiveness of our proposed joint framework for real DP images.
The qualitative comparisons in Fig.~\ref{fig:real_deblur_vis} further show that CrossDD produces visually cleaner and sharper restoration results than the competing methods on real-scene DP images.
CrossDD also consistently outperforms the joint-task method DDDNet on both tasks on the real DP5K-Disp dataset.

More visual comparisons on additional synthetic and real-world scenes can be seen in the supplementary material.

\begin{table}[t]
\centering
\caption{Ablation study on disparity guidance and image guidance. All variants use the same number of recurrent updates and residual blocks, while the corresponding guidance input is removed when disabled. The best results are highlighted in bold.}
\vspace{-2mm}
\label{tab:ablation_main}
\scriptsize
\setlength{\tabcolsep}{2.5pt}
\renewcommand{\arraystretch}{1.05}
\resizebox{\columnwidth}{!}{%
\begin{tabular}{cc|ccccc|ccccc}
\toprule
& 
& \multicolumn{5}{c|}{QPD-Data}
& \multicolumn{5}{c}{DP5K-Disp} \\
\cmidrule(lr){3-7} 
\cmidrule(lr){8-12}

& 
& \multicolumn{2}{c}{Disparity}
& \multicolumn{3}{c|}{Deblur}
& \multicolumn{2}{c}{Disparity}
& \multicolumn{3}{c}{Deblur} \\
\cmidrule(lr){3-4}
\cmidrule(lr){5-7} 
\cmidrule(lr){8-9}
\cmidrule(lr){10-12}

Disp. Guide & Image Guide
& EPE$\downarrow$ & RMSE$\downarrow$
& PSNR$\uparrow$ & SSIM$\uparrow$ & MAE$\downarrow$
& EPE$\downarrow$ & RMSE$\downarrow$
& PSNR$\uparrow$ & SSIM$\uparrow$ & MAE$\downarrow$ \\
\midrule

\xmark & \xmark
& 0.0258 & 0.0741
& 35.478 & 0.9688 & 0.0091
& 0.0679 & 0.1106
& 34.107 & 0.9525 & 0.0133 \\

\xmark & \cmark
& 0.0241 & 0.0696
& 35.584 & 0.9635 & 0.0089
& 0.0662 & 0.1049
& 34.257 & 0.9491 & 0.0133 \\

\cmark & \xmark
& 0.0252 & 0.0720
& 36.922 & 0.9732 & 0.0079
& 0.0678 & 0.1101
& 34.663 & 0.9544 & 0.0128 \\

\cmark & \cmark
& {\bfseries 0.0232} & {\bfseries 0.0675}
& {\bfseries 37.234} & {\bfseries 0.9737} & {\bfseries 0.0075}
& {\bfseries 0.0643} & {\bfseries 0.1045}
& {\bfseries 34.851} & {\bfseries 0.9552} & {\bfseries 0.0124} \\
\bottomrule
\end{tabular}%
}
\end{table}

\begin{figure*}[t]
    \centering
    \includegraphics[width=\linewidth]{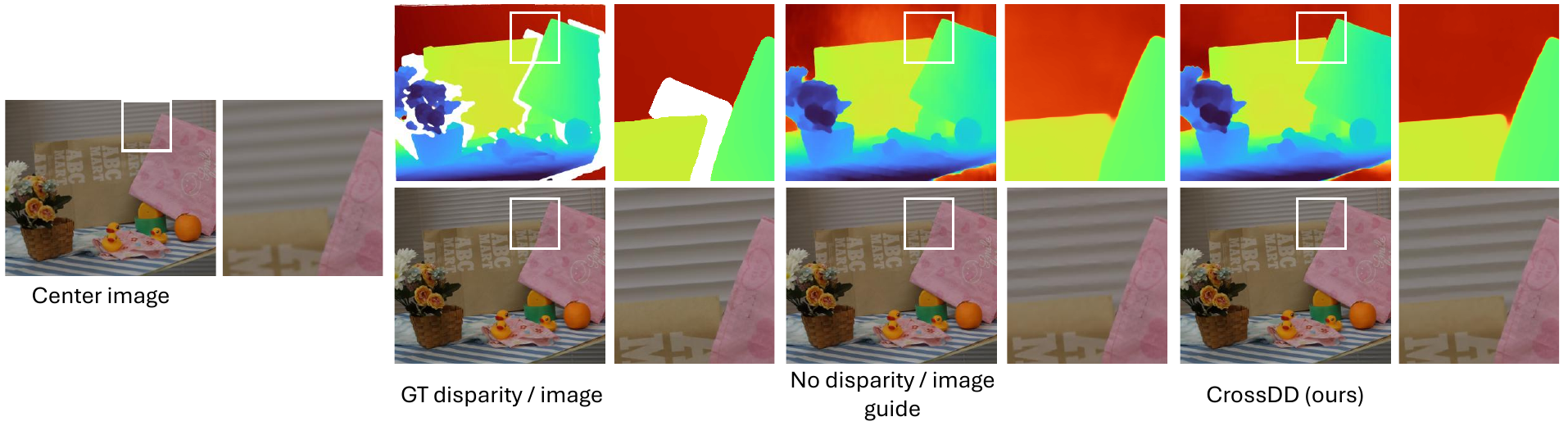}\\
    \vspace{-3mm}
    \caption{
    Qualitative results of the ablation study on DP5K-Disp. 
    Compared with the baseline without disparity and image guidance, CrossDD produces cleaner disparity maps with sharper boundaries and restores sharper image details.
    }
    \label{fig:ablation_vis}
    \vspace{-3mm}
\end{figure*}

\subsection{Ablation Studies}

We conducted ablation experiments to analyze the effects of disparity guidance in DGIR and image guidance in IGDR. All variants retain the same recurrent refinement architecture, the same number of GRU updates, and the same number of residual blocks. When disparity guidance is disabled, DGIR refines the current image without taking the disparity map as input. Similarly, when image guidance is disabled, IGDR refines the current disparity without taking the current image as input. Thus, the model capacities of all variants are comparable, and the main difference lies in whether cross-task guidance is provided.

The quantitative results are summarized in Table~\ref{tab:ablation_main}, and Fig.~\ref{fig:ablation_vis} shows qualitative comparisons on DP5K-Disp.
First, enabling disparity guidance consistently improves defocus deblurring on both QPD-Data and DP5K-Disp compared with the variant without guidance. This result shows that the disparity map provides useful spatially varying geometry cues for defocus deblurring.
Next, enabling image guidance improves disparity estimation on both datasets, confirming that image structures are useful for refining disparity estimates. However, the deblurring improvement is limited in this setting because the image refinement stream does not use disparity guidance.
Finally, enabling both disparity and image guidance achieves the best performance on all metrics. Compared with the variants using only one type of guidance, the full model further improves both disparity estimation and defocus deblurring on both datasets. These results demonstrate that disparity-guided image refinement and image-guided disparity refinement are complementary, and their interaction benefits both tasks.

Additional ablation studies are reported in the supplementary material to further validate the proposed design, including the effect of enabling cross-task guidance across different numbers of CRR modules and the effectiveness of the two-stage training strategy.

\section{Limitations}

Although CrossDD achieves strong performance in both disparity estimation and defocus deblurring, it still has several limitations.
First, the framework prioritizes performance over efficiency. It combines a recurrent disparity-estimation backbone with CRR modules after each recurrent update, and each DGIR/IGDR module contains multiple residual blocks. Moreover, since DGIR relies on intermediate disparity estimates, we adopt a two-stage training strategy to stabilize optimization. These design choices improve performance, but increase training cost and model complexity.
Second, the current real-world benchmark is constructed using only the F4.0 setting of DP5K, which corresponds to moderate blur levels. Therefore, the generalization ability of CrossDD to more diverse aperture settings and stronger blur conditions remains to be further validated.

\section{Conclusion}

In this paper, we have proposed CrossDD, a unified recurrent network architecture for joint DP disparity estimation and defocus deblurring. CrossDD is featured by a CRR module, which performs cross-task refinement after each recurrent update.
Specifically, the DP disparity and the deblurred image are used as guides to refine each other. 
To train and evaluate CrossDD on real-world data, we have presented a depth-to-disparity conversion process to repurpose the DP5K dataset for our joint task. The converted dataset, namely DP5K-Disp, serves as a practical benchmark for future research. 
Experimental results have demonstrated the effectiveness of CrossDD, which achieves strong performance on both disparity estimation and defocus deblurring, validating that iterative cross-task interaction is beneficial for exploiting complementary information observed in DP image sensing.
In future work, it would be interesting to reduce the computational cost of the current recurrent network while maintaining its performance. Another important direction is to validate the generalization ability of the method under more diverse real-world defocus blur conditions, beyond the current single-aperture setting.
In addition, building higher-quality real-world benchmarks with better sharp/blur consistency and more accurate disparity ground truths would help further improve the joint task of DP disparity estimation and defocus deblurring.

\clearpage
\section*{Supplementary Material}
\setcounter{section}{0}
\setcounter{figure}{0}
\setcounter{table}{0}
\setcounter{equation}{0}
\renewcommand{\thesection}{S\arabic{section}}
\renewcommand{\thefigure}{S\arabic{figure}}
\renewcommand{\thetable}{S\arabic{table}}
\renewcommand{\theequation}{S\arabic{equation}}
\renewcommand{\theHsection}{supp.\arabic{section}}
\renewcommand{\theHfigure}{supp.\arabic{figure}}
\renewcommand{\theHtable}{supp.\arabic{table}}
\renewcommand{\theHequation}{supp.\arabic{equation}}

\section{Details of Real-World Dataset Construction}
\label{supp:dataset_details}

This section provides additional details of the real-world dataset construction described in Sec~.4 of the main paper.
Specifically, we describe the photometric alignment used to reduce brightness and vignetting inconsistencies between DP images and sharp images, and the depth-to-disparity conversion used to construct the DP5K-Disp dataset from the original DP5K depth maps.

\subsection{Photometric Alignment}

\begin{figure}[t]
\begin{center}
\centerline{\includegraphics[width=\linewidth]{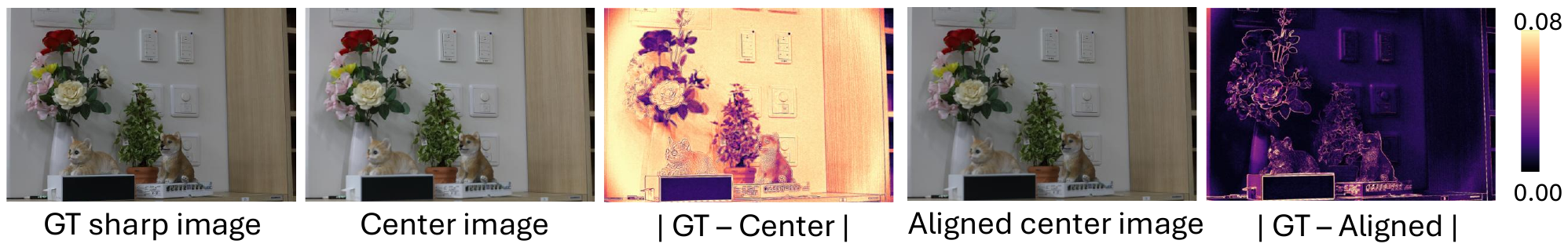}}
\end{center}
   \vspace{-10mm}
    \caption{Example of photometric alignment. From left to right: ground-truth sharp image, original center image, absolute difference before alignment, aligned center image, and absolute difference after alignment. Photometric alignment reduces the intensity discrepancy between the sharp and blurred images.}
\label{fig:alignment_examples}
\end{figure}

We constructed our DP5K-Disp dataset based on DP5K~\cite{li2023learning}, an existing real-world DP dataset. In DP5K, DP image pairs are captured under several F-number settings, while the sharp images used as deblurring ground truths are captured at F22 with a large depth of field.
We observed that this causes brightness inconsistencies because the camera may automatically adjust image brightness during image capture. In addition, the DP image pairs captured at small F-numbers exhibit noticeable vignetting due to the optical characteristics of the DSLR camera. These photometric inconsistencies between the input DP images and the ground-truth sharp images, which can be observed in the first two columns of Fig.~\ref{fig:alignment_examples}, significantly disturb network training. Thus, we corrected these inconsistencies in a simple photometric alignment manner.

We first aligned the brightness of the DP images to that of the ground-truth sharp image by applying a linear scaling as
\begin{equation}
\hat{\mathbf{I}}_X = a\mathbf{I}_X,
\end{equation}
where $\mathbf{I}_X, \ X \in \{L, C, R\}$ represents each of the left, the center, and the right images in the DP images. The center image was first generated by averaging the left and right DP images as
$\mathbf{I}_C=(\mathbf{I}_L+\mathbf{I}_R)/2$. The scale coefficient $a$ was estimated using the center image as
\begin{equation}
a = \arg\min_a||\mathbf{I}_S - a\mathbf{I}_C||_2,
\end{equation}
where $\mathbf{I}_S$ is the sharp image. Here, we only used a $200 \times 200$ image crop at the image center to discount the effect of vignetting for this global brightness alignment.

Then, we performed vignetting correction. 
Following the radial modeling of Goldman and Chen~\cite{goldman2010vignette}, which models vignetting as an even polynomial of the distance from the image center, we adopted a simplified 4th-order formulation as
\begin{equation}
I_S \approx \left( 1 + k_1 r^2 + k_2 r^4 \right) \hat{I}_X,
\end{equation}
where $I_S$ denotes a pixel value in the sharp image, $\hat{I}_X$ denotes the corresponding pixel value of the brightness-aligned DP images, $r$ is the normalized distance from the image center, and $k_1$ and $k_2$ are the vignetting coefficients.
To estimate $k_1$ and $k_2$ robustly, we used all brightness-aligned center images generated from DP pairs captured with the same F-number setting and fitted the model by least squares as
\begin{equation}
(k_1, k_2) = \arg\min_{k1,k2} \sum_i  ||I_S^i - \left( 1 + k_1 r^2 + k_2 r^4 \right) \hat{I}^i_C||_2,
\end{equation}
where $i$ represents a sampled pixel from the dataset images. Since vignetting is a radial effect, we sampled pixels with even probability according to the distance from the image center, $r$, for reliable estimation.

After estimating $k_1$ and $k_2$, the vignetting correction was applied to the brightness-aligned left and right DP images as
\begin{equation}
I'_X = \left(1+k_1r^2+k_2r^4\right)\hat{I}_X,
\quad X\in\{L,R\}.
\end{equation}
where $I'_X$ denotes the vignetting-corrected DP image.
The final corrected center image was then generated by averaging the corrected left and right images as
\begin{equation}
I'_C=\frac{I'_L+I'_R}{2}.
\end{equation}
Therefore, the photometric relationship between the center image and the left and right DP images is preserved in the processed data.

As shown in Fig.~\ref{fig:alignment_examples}, the photometric alignment reduces the discrepancy between the ground-truth sharp image and the input DP image.

\subsection{Depth-to-Disparity Conversion}
\label{supp:depth_to_disparity}

We adopted the thin-lens DP disparity model~\cite{Kim2023Dualpixel}.
In this model, DP disparity is proportional to the circle of confusion and is expressed by a function of the focus distance and scene depth as
\begin{equation}
\begin{aligned}
D
&= \alpha \frac{Af}{1-f/f_d}
\left(\frac{1}{f_d}-\frac{1}{d}\right) \\
&= \alpha Af \frac{d-f_d}{d(f_d-f)} ,
\end{aligned}
\end{equation}
where $D$ denotes the signed DP disparity, $A$ is the aperture diameter, $f$ is the focal length, $f_d$ is the focus distance, $d$ is the depth value, and $\alpha$ is an ambiguous scale coefficient.
Here, $A$, $f$, and $\alpha$ are camera-dependent quantities, whereas $d$ and $f_d$ depend on the scene geometry and focus setting.

Since the focal length $f$ is much smaller than the focus distance $f_d$ in our setting, we approximate $f_d-f \approx f_d$.
We further absorb the camera-dependent terms and other fixed optical factors into a single blur scale parameter $K$, obtaining
\begin{equation}
D = K\frac{d-f_d}{d \cdot f_d}.
\end{equation}

The blur scale parameter $K$ is not provided in DP5K and is difficult to measure directly.
Therefore, we estimated it using photometric consistency between the DP views.
Given a candidate $K$, we first converted the depth map into a disparity map using the above equation and the focus distance provided by the dataset.
We then warped the left and right DP images toward the center view using the converted disparity, producing warped left-to-center and right-to-center images.
The quality of the converted disparity was evaluated by the photometric consistency between these two warped images.
Specifically, we minimized the photometric $L_2$ loss between the warped left-to-center and right-to-center images as
\begin{equation}
K =
\arg\min_K \sum_j \left\|
W(\mathbf{I}^j_L, \mathbf{D}^j(K)) -
W(\mathbf{I}^j_R, -\mathbf{D}^j(K))
\right\|_2 ,
\end{equation}
where $W(\cdot)$ denotes image warping, $\mathbf{D}(K)$ is the disparity map converted from the depth map using the candidate $K$, and $j$ is the index for database images.
We used a bounded Brent method~\cite{brent2013algorithms} to optimize $K$ over the range $[0,100]$.
Since this study uses the F4.0 setting for input DP image pairs, all F4.0 images were used for estimating $K$ and constructing DP5K-Disp.

\section{Additional Analysis of CrossDD}

This section provides additional analysis of the proposed CrossDD.
In addition to the ablation study in Sec.~5.3 of the main paper, we further examine the effect of the number of CRR modules and the effect of the two-stage training strategy. We also provide additional qualitative results for the ablation study to support the quantitative observations.

\subsection{Effect of Cross-Task Guidance Across Recurrent Updates}

We investigate the effect of the number of CRR modules with enabled cross-task guidance. All settings use eight CRR modules with identical residual refinement architectures. The difference is the number of CRR modules in which disparity guidance in DGIR and image guidance in IGDR are enabled.

Specifically, when $N_{\mathrm{guided}}=1$, cross-task guidance is enabled only in the final CRR module. When $N_{\mathrm{guided}}=4$, it is enabled in the final four CRR modules. When $N_{\mathrm{guided}}=8$, cross-task guidance is enabled in all CRR modules.

\begin{table}[t]
\centering
\caption{Effect of the number of CRR modules with cross-task guidance on QPD-Data and DP5K-Disp.
Eight CRR modules with identical architectures are used in all settings.
In the final $N_{\mathrm{guided}}$ CRR modules, DGIR receives disparity guidance and IGDR receives image guidance; in the remaining CRR modules, the corresponding guidance inputs are omitted.
The best results are highlighted in bold.}
\vspace{-2mm}
\label{tab:ablation_iter}
\scriptsize
\setlength{\tabcolsep}{2.5pt}
\renewcommand{\arraystretch}{1.05}
\resizebox{\columnwidth}{!}{%
\begin{tabular}{c|ccccc|ccccc}
\toprule
& \multicolumn{5}{c|}{QPD-Data}
& \multicolumn{5}{c}{DP5K-Disp} \\
\cmidrule(lr){2-6}
\cmidrule(lr){7-11}

$N_{\mathrm{guided}}$
& \multicolumn{2}{c}{Disparity}
& \multicolumn{3}{c|}{Deblur}
& \multicolumn{2}{c}{Disparity}
& \multicolumn{3}{c}{Deblur} \\
\cmidrule(lr){2-3}
\cmidrule(lr){4-6}
\cmidrule(lr){7-8}
\cmidrule(lr){9-11}

& EPE$\downarrow$ & RMSE$\downarrow$
& PSNR$\uparrow$ & SSIM$\uparrow$ & MAE$\downarrow$
& EPE$\downarrow$ & RMSE$\downarrow$
& PSNR$\uparrow$ & SSIM$\uparrow$ & MAE$\downarrow$ \\
\midrule

1
& 0.0242 & 0.0697
& 36.668 & 0.9715 & 0.0081
& 0.0655 & 0.1080
& 34.726 & 0.9547 & 0.0127 \\

4
& 0.0238 & 0.0693
& 36.995 & 0.9732 & 0.0077
& 0.0648 & 0.1066
& 34.769 & 0.9551 & 0.0126 \\

8
& \textbf{0.0232} & \textbf{0.0675}
& \textbf{37.234} & \textbf{0.9737} & \textbf{0.0075}
& \textbf{0.0643} & \textbf{0.1045}
& \textbf{34.851} & \textbf{0.9552} & \textbf{0.0124} \\
\bottomrule
\end{tabular}%
}
\vspace{3mm}
\end{table}

As shown in Table~\ref{tab:ablation_iter}, enabling cross-task guidance in more CRR modules consistently improves both disparity estimation and defocus deblurring on QPD-Data and DP5K-Disp. On QPD-Data, increasing $N_{\mathrm{guided}}$ from 1 to 8 reduces EPE from 0.0242 to 0.0232 and improves PSNR from 36.668 to 37.234. A consistent, although smaller, improvement is also observed on DP5K-Disp. These results indicate that repeatedly exchanging disparity-aware and image-aware guidance throughout the recurrent process is more effective than enabling such guidance only in the final refinement stage.

\subsection{Effect of Two-Stage Training Strategy}

We further validate the effectiveness of the two-stage training strategy described in Sec.~\textcolor{red}{3.3} of the main paper.
The 1-stage baseline trains the full network with all CRR modules enabled from random initialization using the joint loss in Eq.~\textcolor{red}{3} of the main paper.
In contrast, our 2-stage strategy first trains the recurrent disparity estimation backbone without CRR modules, and then jointly optimizes the disparity backbone and all CRR modules.

\begin{table}[t]
\centering
\caption{Ablation study of the two-stage training strategy on QPD-Data and DP5K-Disp.
The 1-stage baseline trains the full network directly from random initialization, while the 2-stage strategy first pretrains the disparity backbone and then enables joint optimization with CRR modules.
The best results are highlighted in bold.}
\vspace{-2mm}
\label{tab:ablation_two_stage}
\scriptsize
\setlength{\tabcolsep}{2.5pt}
\renewcommand{\arraystretch}{1.05}
\resizebox{\columnwidth}{!}{%
\begin{tabular}{c|ccccc|ccccc}
\toprule
& \multicolumn{5}{c|}{QPD-Data}
& \multicolumn{5}{c}{DP5K-Disp}
\tabularnewline

\cmidrule(lr){2-6}
\cmidrule(lr){7-11}

Strategy
& \multicolumn{2}{c}{Disparity}
& \multicolumn{3}{c|}{Deblur}
& \multicolumn{2}{c}{Disparity}
& \multicolumn{3}{c}{Deblur}
\tabularnewline

\cmidrule(lr){2-3}
\cmidrule(lr){4-6}
\cmidrule(lr){7-8}
\cmidrule(lr){9-11}

& EPE$\downarrow$ & RMSE$\downarrow$
& PSNR$\uparrow$ & SSIM$\uparrow$ & MAE$\downarrow$
& EPE$\downarrow$ & RMSE$\downarrow$
& PSNR$\uparrow$ & SSIM$\uparrow$ & MAE$\downarrow$
\tabularnewline

\midrule

1-stage
& 0.0253 & 0.0730
& 37.104 & 0.9736 & 0.0076
& 0.0657 & 0.1064
& 34.812 & 0.9551 & 0.0125
\tabularnewline

2-stage (Ours)
& \textbf{0.0232} & \textbf{0.0675}
& \textbf{37.234} & \textbf{0.9737} & \textbf{0.0075}
& \textbf{0.0643} & \textbf{0.1045}
& \textbf{34.851} & \textbf{0.9552} & \textbf{0.0124}
\tabularnewline

\bottomrule
\end{tabular}%
}
\vspace{3mm}
\end{table}

\begin{figure}[t!]
\vspace{3mm}
\centering
\includegraphics[width=\textwidth]{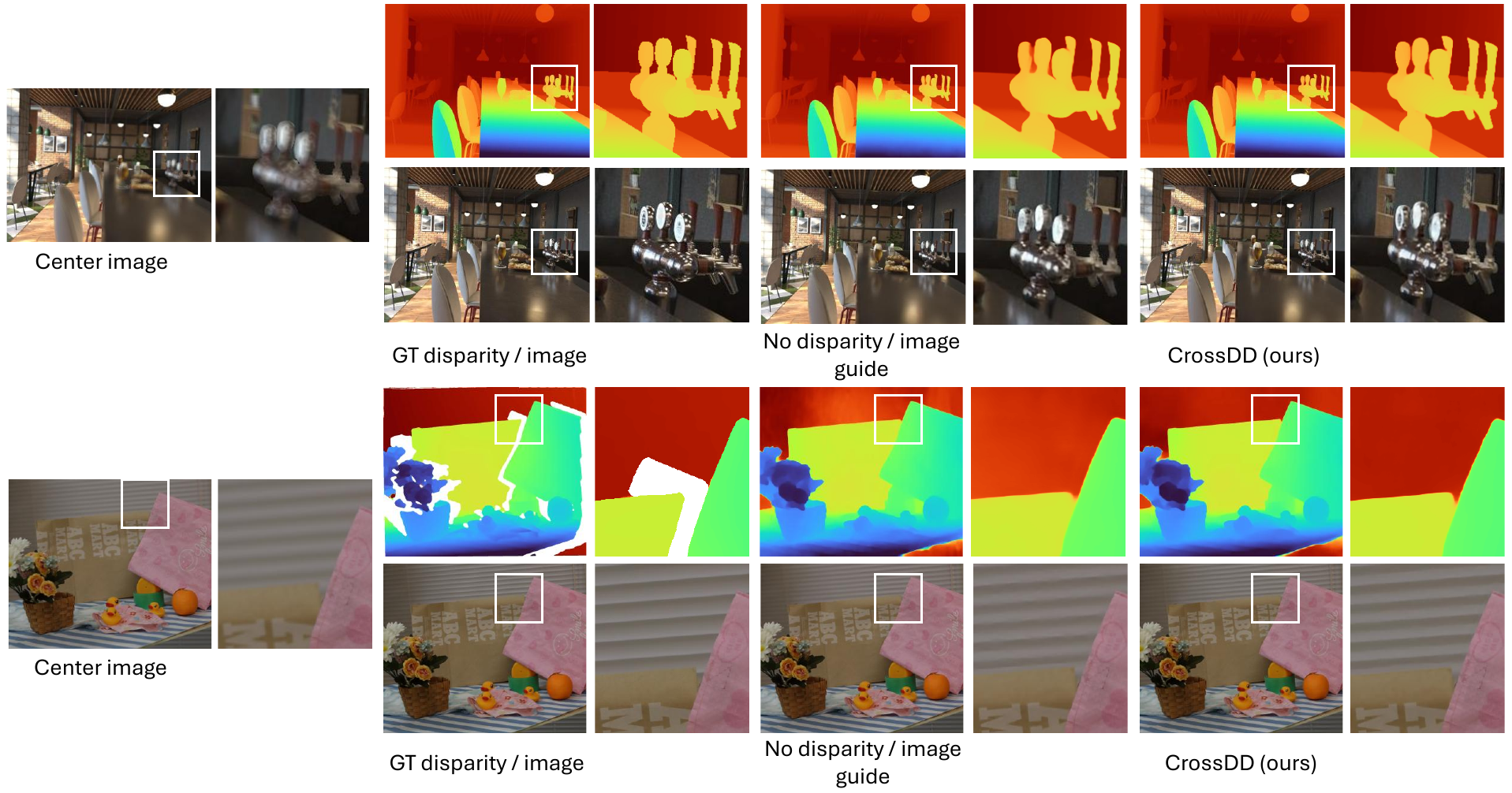}
\vspace{-6mm}
\caption{
Qualitative results for the ablation study on QPD-Data (top two rows) and DP5K-Disp (bottom two rows).
Compared with the variant that does not use disparity and image guidance, CrossDD produces cleaner disparity maps with sharper boundaries and restores sharper image details.
}
\label{fig:supp_ablation_qpd}
\vspace{-3mm}
\end{figure}

\begin{figure}[t]
\vspace{3mm}
\centering
\includegraphics[width=\textwidth]{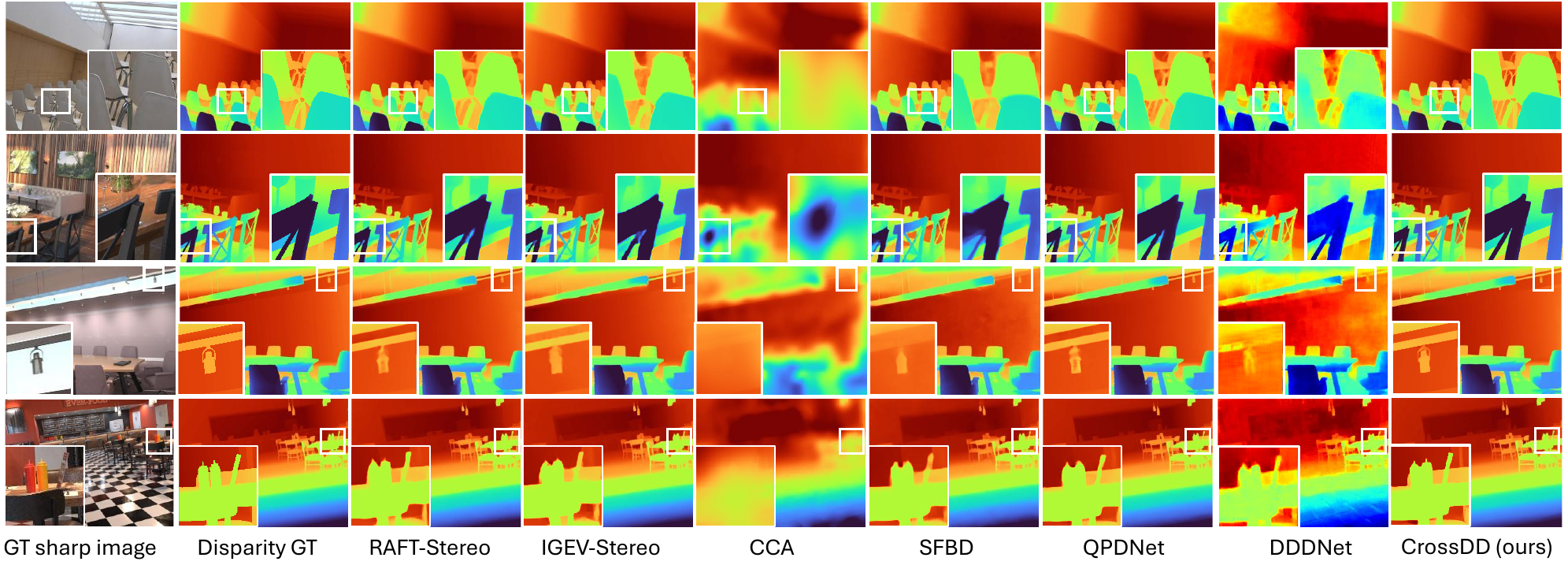}
\vspace{-6mm}
\caption{
Qualitative comparison of disparity estimation on the synthetic QPD-Data.
}
\label{fig:supp_qpd_disp}
\vspace{3mm}
\end{figure}

\begin{figure}[t]
\centering
\includegraphics[width=1.0\textwidth]{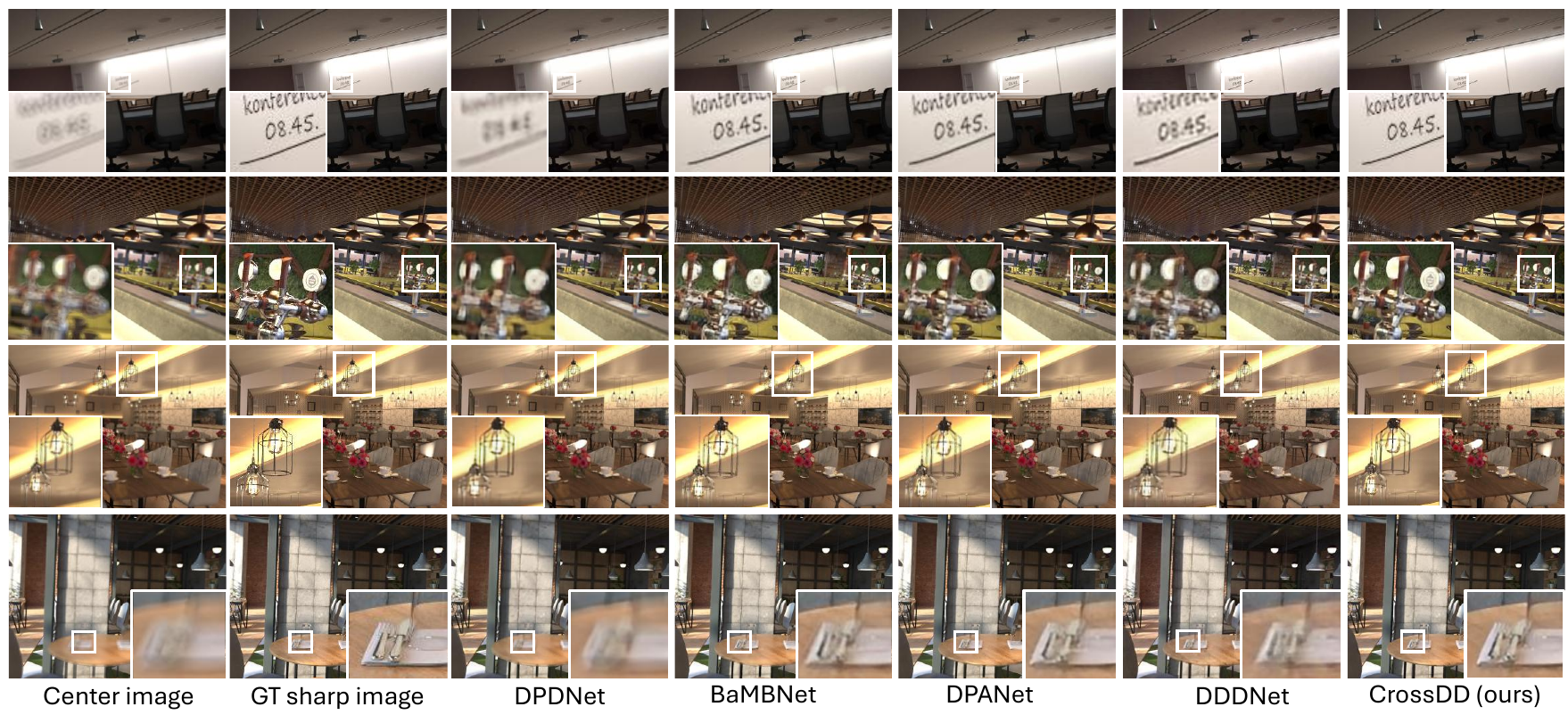}
\vspace{-6mm}
\caption{
Qualitative comparison of defocus deblurring on the synthetic QPD-Data.
}
\label{fig:supp_qpd_deblur}
\vspace{3mm}
\end{figure}

As shown in Table~\ref{tab:ablation_two_stage}, the proposed 2-stage training strategy consistently outperforms direct 1-stage training on both datasets.
The improvement is observed in both disparity estimation and defocus deblurring metrics, indicating that pretraining the disparity backbone provides more reliable geometric estimates before enabling cross-reference refinement.
This suggests that directly optimizing the full joint framework from random initialization can make the early cross-task guidance less stable, whereas the 2-stage strategy provides a more stable initialization for joint disparity and image refinement.

\subsection{Additional Visualization Results for Ablation Study}
\label{supp:ablation_vis}

The top of Fig.~\ref{fig:supp_ablation_qpd} shows the qualitative results for the ablation study on QPD-Data. In the main paper, we show the visualization results on the real DP5K-Disp dataset at the bottom of Fig.~\ref{fig:supp_ablation_qpd}. Here, we also provide the corresponding synthetic results. Compared with the cases without disparity and image guidance, CrossDD produces cleaner disparity maps with sharper boundaries and restores sharper image details, which is consistent with the quantitative results reported in the main paper.

\section{Additional Qualitative Results}
\label{supp:qualitative}

We provide additional qualitative comparisons on the synthetic QPD-Data and real DP5K-Disp datasets, which cannot be included in the main paper due to space limitations.

\subsection{Synthetic QPD-Data}

Figure~\ref{fig:supp_qpd_disp} shows additional qualitative results of disparity estimation on the synthetic QPD-Data. CrossDD produces cleaner disparity maps with sharper boundaries and more reliable local structures than the compared methods.

Figure~\ref{fig:supp_qpd_deblur} shows additional qualitative results of defocus deblurring on QPD-Data. CrossDD restores clearer textures and sharper local details than the compared methods.

\subsection{Real DP5K-Disp}

\begin{figure}[!t]
\centering
\includegraphics[width=\textwidth]{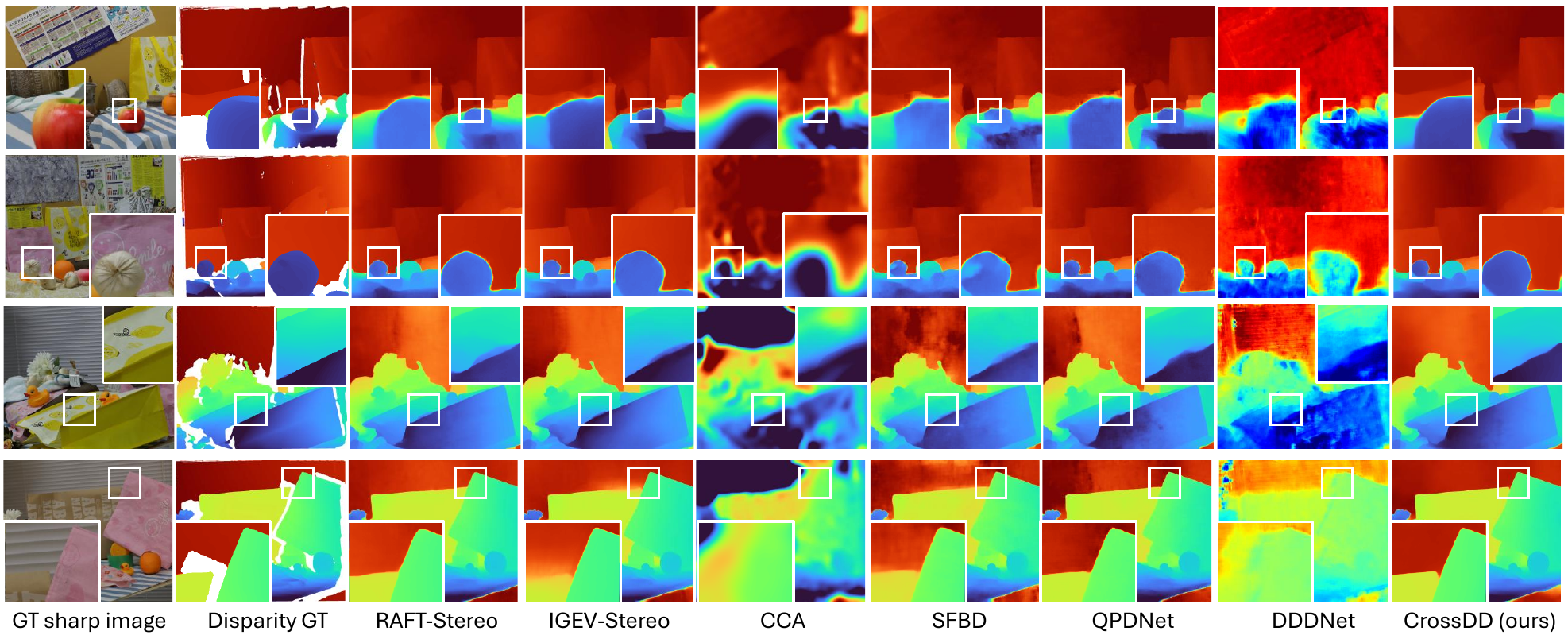}
\vspace{-6mm}
\caption{
Qualitative comparison of disparity estimation on the real DP5K-Disp.
}
\label{fig:supp_dp5k_disp}
\end{figure}

Figure~\ref{fig:supp_dp5k_disp} shows additional qualitative results of disparity estimation on the real DP5K-Disp. CrossDD produces cleaner disparity maps and better preserves object boundaries than the compared methods.

\begin{figure}[!t]
\vspace{3mm}
\centering
\includegraphics[width=\textwidth]{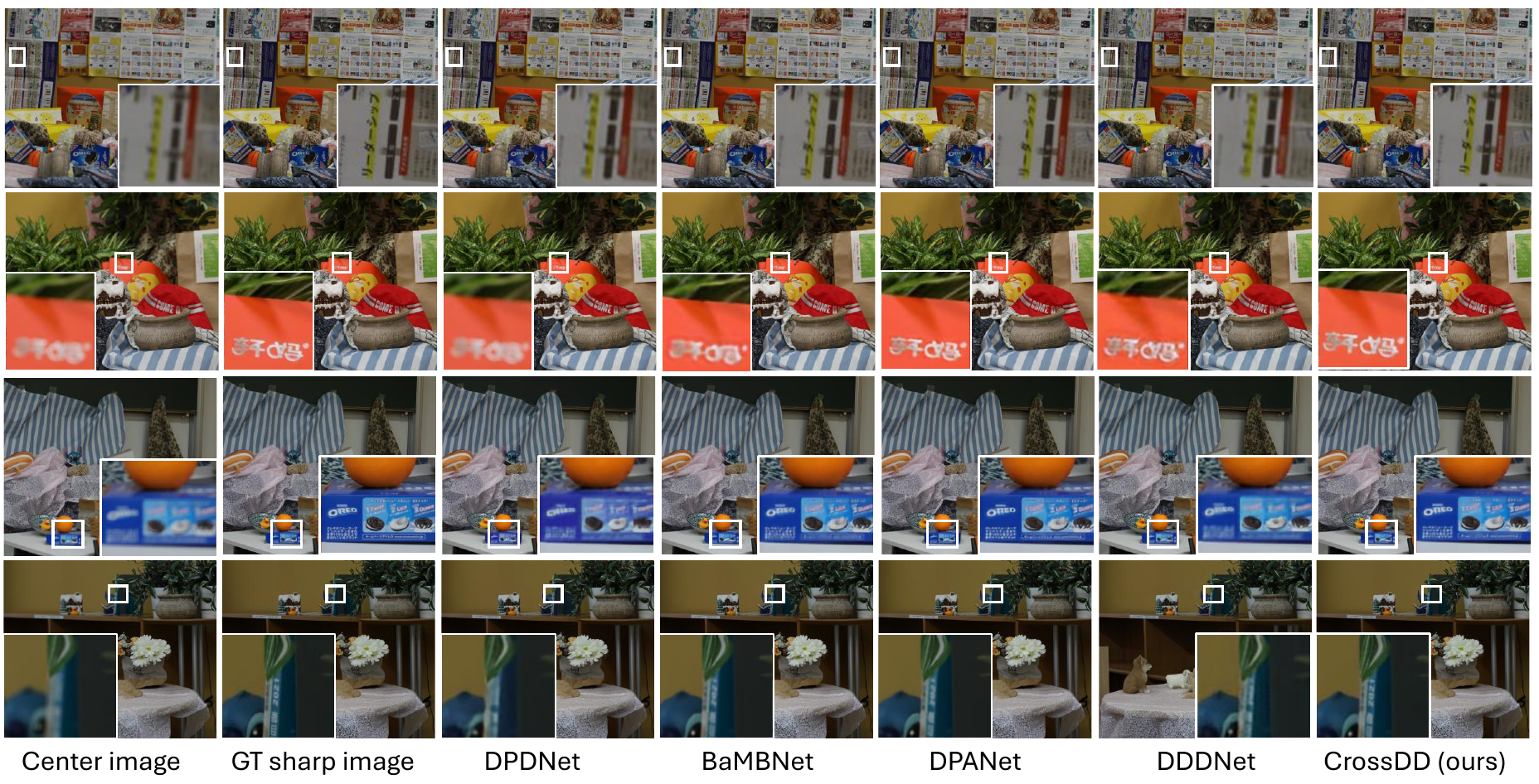}
\vspace{-6mm}
\caption{
Qualitative comparison of defocus deblurring on the real DP5K-Disp.
}
\label{fig:supp_dp5k_deblur}
\vspace{2mm}
\end{figure}

Figure~\ref{fig:supp_dp5k_deblur} shows additional qualitative results of defocus deblurring on DP5K-Disp. CrossDD produces visually clearer restoration results with reduced residual blur artifacts.

\end{document}